\RequirePackage{fix-cm}
\documentclass[smallextended]{svjour3}       
\smartqed  
\usepackage{multirow}
\usepackage{array}
\usepackage{amssymb}
\usepackage{graphicx}
\usepackage{caption}
\usepackage{float}
\usepackage{subfigure}
\usepackage{amsmath} 

\begin{document}

\title{JOINT EMBEDDING LEARNING OF EDUCATIONAL KNOWLEDGE GRAPHS
}

\titlerunning{JOINT EMBEDDING LEARNING}

\author{Siyu Yao         \and
        Ruijie Wang, Shen Sun, Derui Bu, Jun Liu 
}


\institute{Siyu Yao \at
              No.28, Xianning West Road, Xi'an, Shaanxi, 710049, P.R. China \\
              Tel.: +86-18552129377\\
              \email{cherylmissing@gmail.com}           
}

\date{}

\maketitle

\begin{abstract}
As an efficient model for knowledge organization, the knowledge graph has been widely adopted in several fields, e.g., biomedicine, sociology, and education. And there is a steady trend of learning embedding representations of knowledge graphs to facilitate knowledge graph construction and downstream tasks. In general, knowledge graph embedding techniques aim to learn vectorized representations which preserve the structural information of the graph. And conventional embedding learning models rely on structural relationships among entities and relations. However, in educational knowledge graphs, structural relationships are not the focus. Instead, rich literals of the graphs are more valuable. In this paper, we focus on this problem and propose a novel model for embedding learning of educational knowledge graphs. Our model considers both structural and literal information and jointly learns embedding representations. Three experimental graphs were constructed based on an educational knowledge graph which has been applied in real-world teaching. We conducted two experiments on the three graphs and other common benchmark graphs. The experimental results proved the effectiveness of our model and its superiority over other baselines when processing educational knowledge graphs.
\end{abstract}

\section{INTRODUCTION}
In recent years, besides encyclopedia knowledge graphs (KGs), e.g., Freebase~\cite{bollacker2008freebase}, Yago~\cite{suchanek2007yago}, and DBpedia~\cite{auer2007dbpedia}, a variety of domain-specific KGs~\cite{zheng2019knowledgeforest,messina2017biograkn}, including educational KGs (e.g., Knowledge Forest~\cite{zheng2019knowledgeforest} and KnowEdu~\cite{chen2018knowedu}), have been constructed. Educational KGs organize the knowledge related to teaching in the form of knowledge graphs. An educational KG includes a set of triples, e.g., (\emph{Data\_Structure\_Course, topic, Hash\_Table}), which consist of head entities (e.g., \emph{Data\_Structure\_Course}), relations (e.g., \emph{topic}), and tail entities (e.g., \emph{Hash\_Table}). The entities may represent courses (e.g., \emph{Data\_Structure\_Course}), topics (e.g., \emph{Hash\_Table}), and literals (e.g., the definition of Hash\_Table). Compared to other types of KGs, a remarkable feature of educational KGs is the richness of literals. And the literals contain more valuable information than the structural relationships among entities and relations do. For example, in the graph which organizes the knowledge of the data structure course, the literal definition of \emph{Hash\_Table} is more important than the relationship between \emph{Hash\_Table} and \emph{Data\_Structure\_Course}, i.e., \emph{topic}.

KG embedding techniques have drawn a surge of interest in both academia and industry due to the outperforming performances of KG embedding-based methods over KG related tasks, e.g., KG completion~\cite{lin2015learning}, KG-based question answering~\cite{huang2019knowledge}, and KG-based recommendation~\cite{burke2000knowledge}. Generally, KG embedding models learn low-dimensional embedding vectors of entities and relations of KGs, which are regarded as representations of KGs in embedding spaces and can be utilized in downstream tasks without further modification. In embedding spaces, TransE~\cite{bordes2013translating} and its extensions~\cite{wang2014knowledge,lin2015learning,ji2015knowledge} represent a triple as a translation from the head entity to the tail entity through the relation. More sophisticated embedding models~\cite{kazemi2018simple,guo2018knowledge,guan2019} have been constantly proposed with complex mechanisms for better performances over general KGs, e.g., DBpedia and Freebase. However, as far as we know, there are no existing KG embedding models specially designed for educational KGs which have a biased focus on literal information. And there are following challenges to be tackled when proposing an embedding learning model for educational KGs:

\begin{itemize}
\item[1.] How to appropriately incorporate massive literals attached to different entities of educational KGs into embedding representations? We have noticed a model named LiteralE~\cite{kristiadi2018incorporating} which utilizes literals for embedding learning by a simple portable module. However, LiteralE only considers numerical literals which are apparently a very limited part of literals of educational KGs.

\item[2.] Structural information should not be ignored as well since we expect the learned embedding representations still be applicable for common structural tasks of KG embedding models, e.g., link prediction~\cite{bordes2013translating}.

\item[3.] There are no available benchmark educational KGs for embedding learning. And current large-scale educational KGs, e.g., Knowledge Forest and KnowEdu, have their individually designed schemas and ontologies which make the construction of experimental graphs more difficult.
\end{itemize}

In this paper, we propose a novel KG embedding model which jointly considers structural information and literal information of educational KGs. For structural information, we employ an existing translation-based embedding model, i.e., TransE, to learn structural embedding vectors of entities and relations. For literal information, a pre-trained language representation model, i.e., BERT~\cite{devlin2018bert}, is adopted to encode the literals into literal embedding vectors. Then, we jointly train three independent GRUs~\cite{cho2014learning} to combine the structural and literal embedding vectors into joint embedding vectors. To summarize, the contributions of this paper are fourfold:

\begin{itemize}
	\item[1.]To the best of our knowledge, this is the first study of educational KG embedding learning, which effectively leverages the rich literals of educational KGs.
	
	\item[2.]We propose a novel method to jointly learn embedding representations based on pre-trained structural and literal embedding vectors.
	
	\item[3.]Three experimental educational KGs were constructed, and they could be adopted as benchmark datasets for the embedding learning task.
	
	\item[4.]Extensive experiments were conducted, the results of which proved that our model outperforms other baselines on educational KGs
	
\end{itemize}

This paper is organized as follows. We introduce the related work in Section 2. In Section 3, we present our model in detail. Experiments and empirical analyses are reported in Section 4. The paper ends with outlining conclusions and future work in Section 5.


\section{RELATED WORK}

In this section, we introduce the related work of two techniques we employed in our model which are KG embedding techniques and literal representation techniques.

\subsection{KG embedding techniques}
KG embedding techniques~\cite{bordes2013translating,kazemi2018simple,guo2018knowledge,guan2019} learn continuous low-dimensional vector representations of KGs. We divide them into two categories according to, except the structural information represented by triples we can directly observe, whether additional information, e.g., literal, ontological, and logical information, is utilized.

Most existing models only consider the structural information of KGs when learning embeddings. Translation-based models, including TransE~\cite{bordes2013translating} and its extensions~\cite{wang2014knowledge,lin2015learning,ji2016knowledge,xiao2015transa} are typical models of this category. TransE firstly proposed the translation mechanism which regards relations as translation operations in embedding spaces. Inspired by TransE, later translation-based models~\cite{wang2014knowledge,lin2015learning,ji2016knowledge,xiao2015transa} proposed more sophisticated translation mechanisms to achieve better performances. Xiao et al.~\cite{xiao2015transa} map each relation of the KG to a non-negative matrix and use an adaptive Markov distance in the loss function of their model. TransH~\cite{wang2014knowledge} learns different representations of one entity under different relations, which is used to alleviate the issue of 1-to-N, N-to-1, N-to-N relations. Other than translation-based models, there are models which employ similarity-based loss functions including RESCAL~\cite{nickel2011three} and its variant~\cite{yang2014embedding,nickel2016holographic,trouillon2016complex,liu2017analogical}, and we also noticed a series of methods using neural networks~\cite{socher2013reasoning,bordes2014semantic,dong2014knowledge,liu2016probabilistic}. RESCAL~\cite{nickel2011three} encodes the interaction between entities by embedding relations into matrices. Socher et al.~\cite{socher2013reasoning} extend RESCAL and use bilinear tensors to link head entities and tail entities.

Additional information, e.g., literal, ontological and logical information, has been leveraged by another kind of embedding models~\cite{socher2013reasoning,wang2017knowledge,kazemi2018simple,guan2019}. NTN~\cite{socher2013reasoning} is the earliest model to integrate text descriptions into KG embedding learning~\cite{wang2017knowledge}, and the representations of entities are initialized by average vectors of words contained in their names. TEKE~\cite{wang2016text} defines context vectors of entities and relations, and combines context vectors into traditional models, e.g., TransE, Xie et al.~\cite{xie2016representation} and Xu et al.~\cite{xu2016knowledge} encode textual literals by convolutional and recurrent neural networks. LiteralE~\cite{kristiadi2018incorporating} replaces the original entity embeddings of conventional loss functions with literal-enriched vectors, which are defined by learnable parametrized functions. SimplE~\cite{kazemi2018simple} incorporates certain types of background knowledge into the model by weight tying. RUGE~\cite{guo2018knowledge} simultaneously learns information from three aspects, i.e., observed triples of KGs, unlabelled triples whose labels are going to be predicted iteratively, and soft rules extracted automatically from the KG. KEC~\cite{guan2019} embeds entities and concepts of entities jointly via a concept graph.

\subsection{Literal representation techniques} 
Language Models (LMs) have been dominant in literal representation tasks, and they can be divided into two categories which are statistical language models~\cite{jelinek1980interpolated,chen1999empirical} and neural network language models~\cite{bahdanau2014neural,verwimp2017character,peters2018deep}.

The statistical language models measure the plausibility of a predicted text by its probability distribution. N-gram-based models~\cite{jelinek1980interpolated,katz1987estimation,chen1999empirical} are mainstreams of this category, which handle the issue of parameters excess by introducing the Markov hypothesis~\cite{bell1974markov} and adopt smoothing~\cite{zhai2017study} to address the sparsity of data. However, it has some defects, e.g., the lack of context-dependency and generalization ability. 

Neural network language models can be further divided into RNN-based LMs~\cite{mikolov2010recurrent,mikolov2011empirical,sundermeyer2012lstm,verwimp2017character,peters2018deep}, cache-based LMs~\cite{soutner2012neural,grave2016improving,huang2014cache}, and attention-based LMs~\cite{bahdanau2014neural,tran2016recurrent,mei2017coherent}. Inspired by the first RNN-based LM~\cite{mikolov2010recurrent,mikolov2011empirical}, the work by Sundermeyer et al.~\cite{sundermeyer2012lstm} leverages LSTM~\cite{hochreiter1997long} to capture context dependences. The cache-based LM proposed by Soutner et al.~\cite{soutner2012neural} matches the new input and historical data in the cache to overcome the length limitation of context dependencies. Although the RNN-based LM uses context to predict words, it overlooks the correlation between words. Bahdanau et al.~\cite{bahdanau2014neural} attempt to combine the language model with the attention mechanism and propose a structure of attention-based LM. This model assigns different weights to each word to select useful words for prediction. On the basis of attention mechanism, there are some competitive LM and word representation methods, such as BERT~\cite{devlin2018bert} and GPT~\cite{radford2018improving},  which are all based on Transformer~\cite{vaswani2017attention}. The main difference between them is that GPT uses the decoder of Transformer, while BERT uses the encoder of Transformer.

\section{THE PROPOSED MODEL}

In this section, we introduce the notations to be used in the following of paper, define the embedding learning problem of educational KGs, and then present our proposed model in detail.

\begin{figure}[H]
	\centering 
	\subfigure[A partial view of Knowledge Forest.] {
		
		\begin{minipage}{5.5cm}
			\centering
			\includegraphics[scale=0.5]{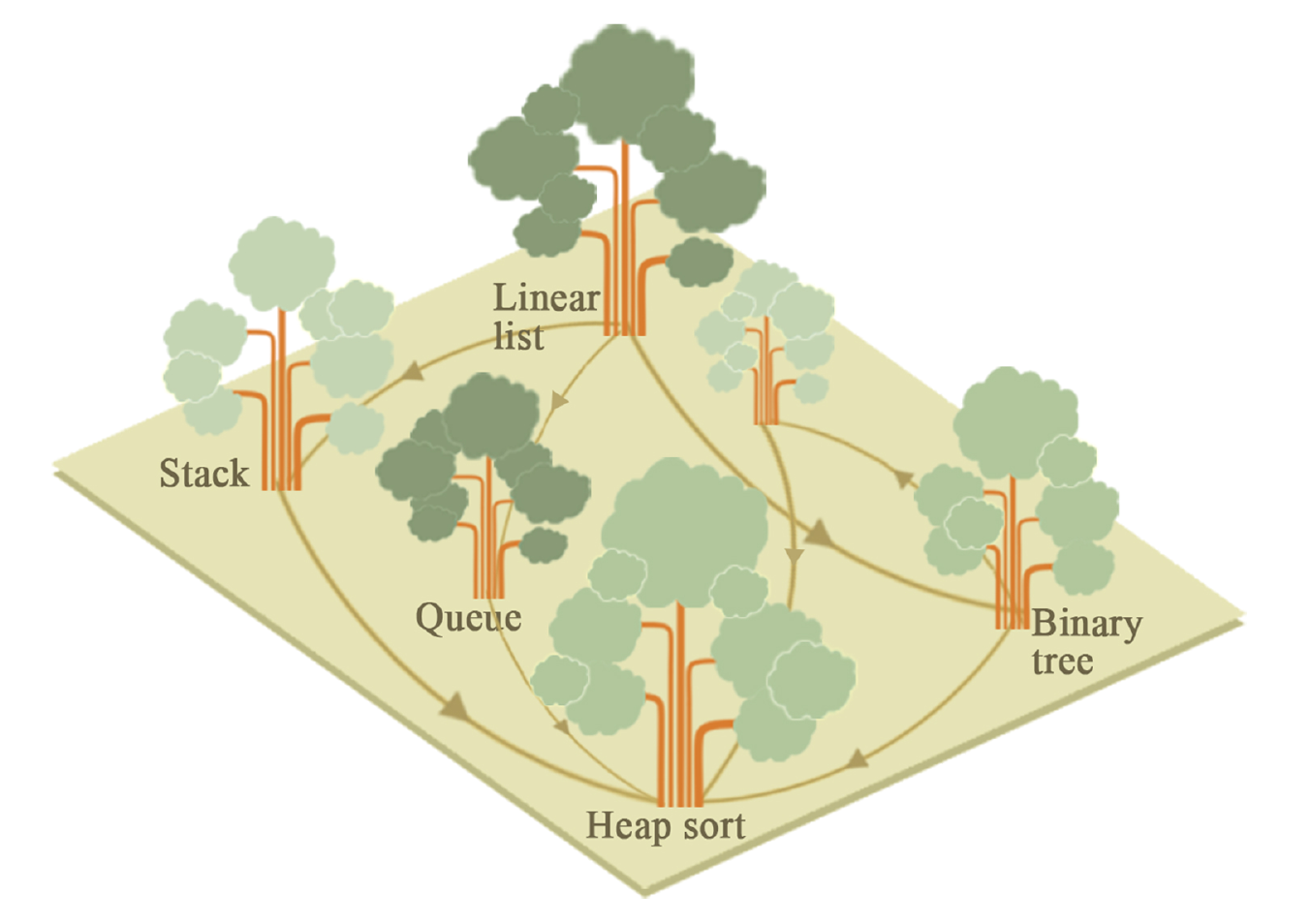}
		\end{minipage}	
		
	}
	\subfigure[A facet tree of the entity \emph{Stack}.] {
		
		\begin{minipage}{5.5cm}
			\centering
			\includegraphics[scale=0.5]{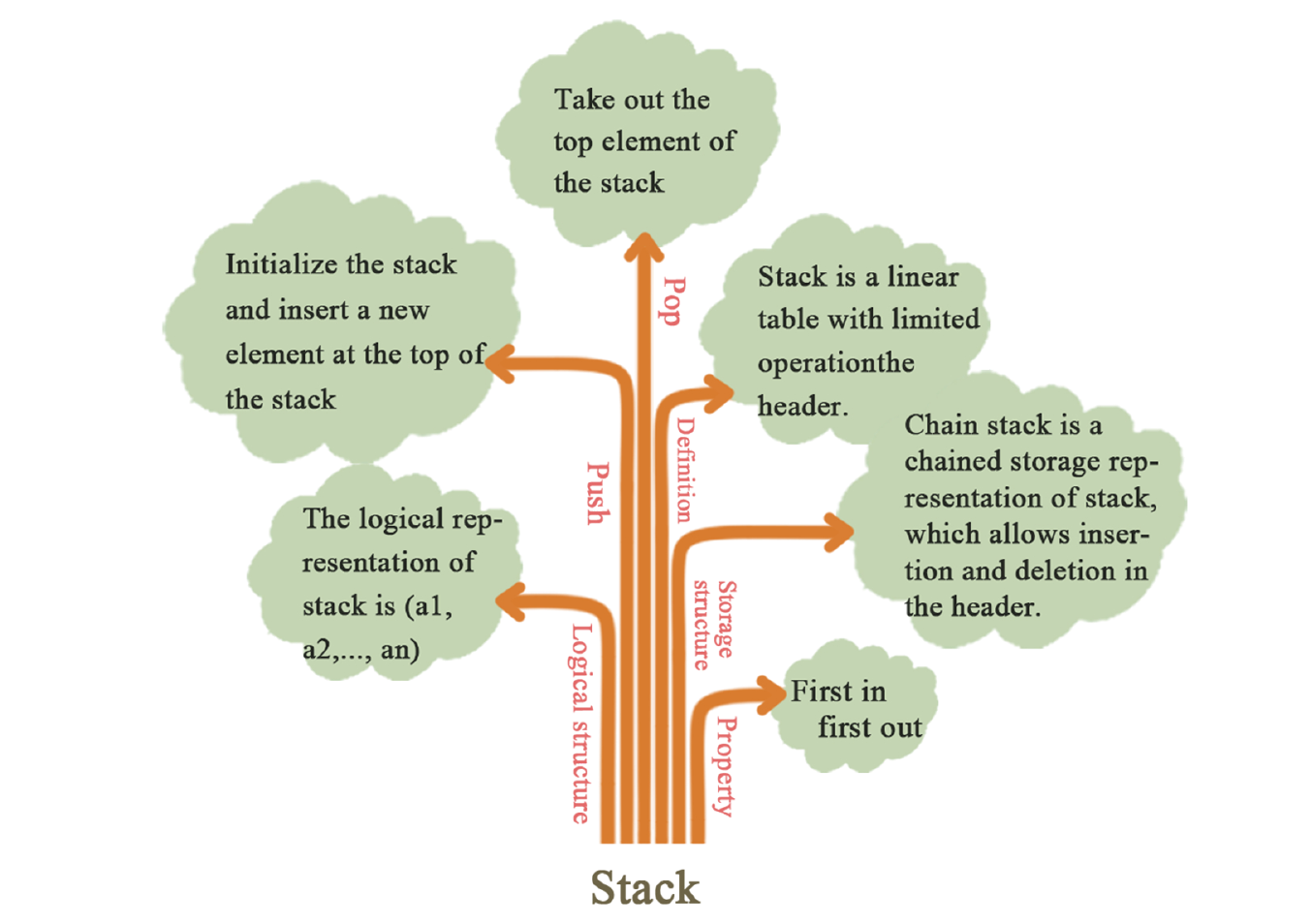}
		\end{minipage}	
		
	}
	
	\caption{A depict of Knowledge Forest}
	\label{fig:1}       
\end{figure}

\subsection{Problem Formulation}

We denote the educational KG as $\mathcal{G} = (\mathcal{E}, \mathcal{R}, \mathcal{L}) $, where $\mathcal{E}$ and $\mathcal{R} $ are sets of entities and relations, respectively, and $\mathcal{L} $ is a set of literals which are attached to the entities and relations. The triple of $\mathcal{G} $ could be denoted as $(h,r,t) \in \mathcal{G}$, where $h,t \in \mathcal{E}$ stand for the head and tail entities, and $r \in \mathcal{R}$ denotes the relation. We define a function $L$ to map the entities and relations to their corresponding literals. Specifically, the literals of $h$, $r$, and $t$ are denoted as $l_h$, $l_r$, and $l_t\in \mathcal{L}$, where $l_h = L(h)$, $l_r = L(r)$, and $l_t = L(t)$. 

The problem of educational KG embedding learning is to learn embedding vectors of entities and relations which capture the structural and literal information of the KG. We employ the translation mechanism of TransE in the joint score function of our model. Therefore, we expect that the learned embedding representations maintain the translation mechanism. And with the literal information integrated, the performance of our model should be significantly improved. Specifically, given a triple $(h,r,t) \in \mathcal{G}$, which consists of two entities $h,t \in \mathcal{E}$, and a relation $r \in \mathcal{R}$, the problem is to learn their embedding vectors $\mathbf{h} $, $\mathbf{t}$, and $\mathbf{r}$, and we expect the equation $\|\mathbf{h} + \mathbf{r} - \mathbf{t}\|^2_2$ holds, which means that in the embedding space, the tail entity should be close to the point computed by translating the head entity through the relation. 

For a better comprehension, here we briefly introduce an educational KG named Knowledge Forest~\cite{zheng2019knowledgeforest}: As illustrated in Fig. \ref{fig:1}(a), Knowledge Forest consists of facet trees, i.e., the green trees named by course topics, and learning dependencies between trees, i.e., the directed paths between facet trees. Each facet tree has several branches linking the topic to several literals, as illustrated in Fig. \ref{fig:1}(b). Firstly, we regard the topics as entities of the KG, and the learning dependencies represent the relations between entities. For example, there is a directed path from topic Linear List to topic Stack, therefore, the KG of Knowledge Forest should include a triple (\emph{Linear List, dependency, Stack}). Secondly, branches of a facet tree of a topic represent the relations between the entity and its related literals. For example, the corresponding literal of the entity Stack includes its definition and properties, as illustrated in Fig. \ref{fig:1}(b).

\begin{figure}[]
	\includegraphics[width=1\textwidth]{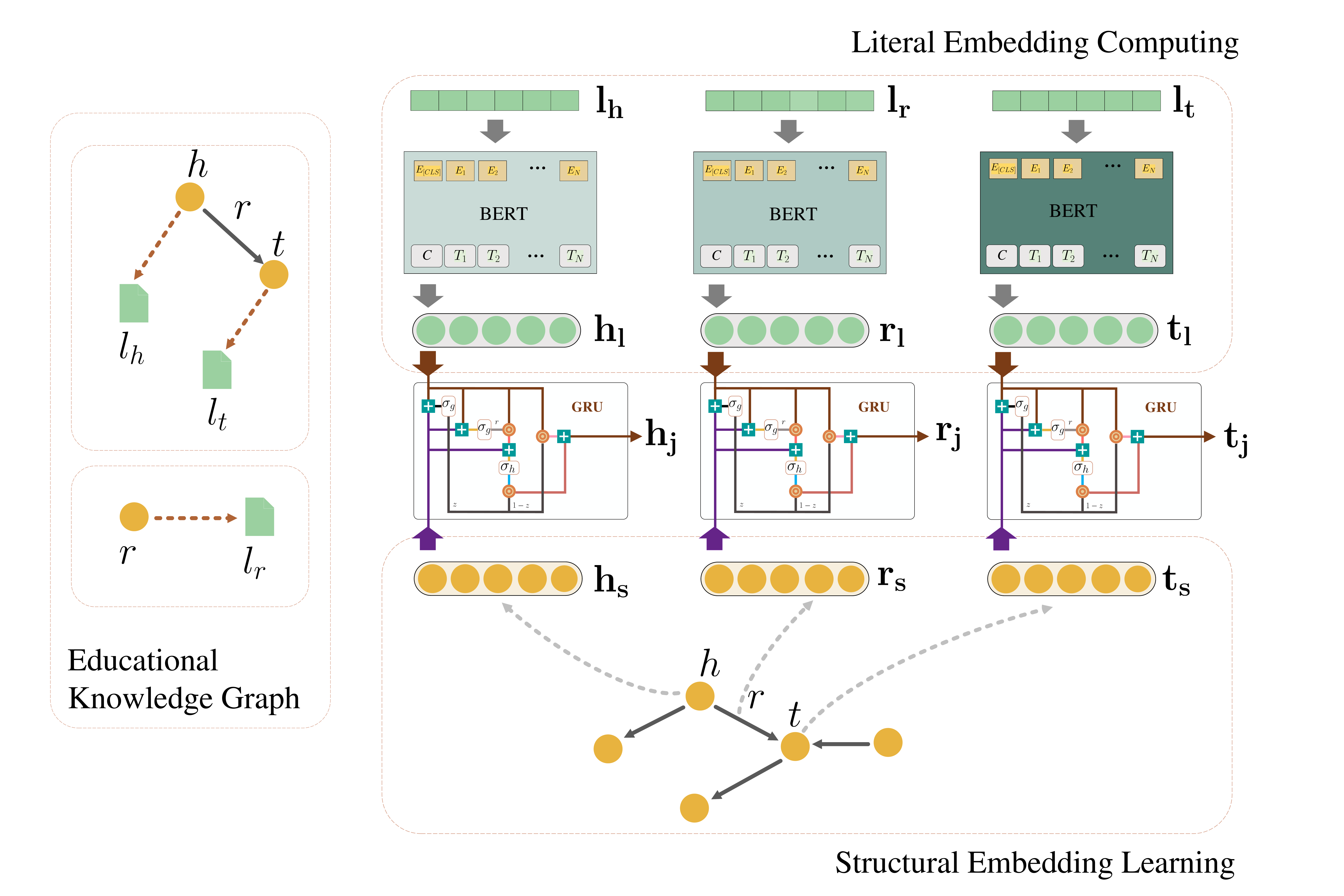}
	\caption{An Overview of Our Model.}
	\label{fig:2}       
\end{figure}

\subsection{Overview of the model}

In this section, we give an overview of our model, as illustrated in Fig. \ref{fig:2}. For an educational KG which consists of triples and literals of the entities and relations, we first learn structural embedding representations based on TransE and literal embedding representations based on BERT~\cite{devlin2018bert}. Then we employ three GRUs to combine the structural and literal embedding representations into joint embedding representations, and adopt the translation mechanism of TransE as the score function of the joint embedding learning to train the GRUs and the joint embedding representations.

\subsection{Structural embedding learning}
The objective of this module is to encode the structural information of the educational KG into structural embedding representations. As we have introduced in Section 2.1, the translation-based models are competitive in capturing the structural relationships among entities and relations. Therefore, we adopt a simple and efficient translation-based model, i.e., TransE, to our structural embedding learning.

Specifically, given an educational KG $\mathcal{G}$, for the triple $(h,r,t) \in \mathcal{G}$, we learn the structural embedding vectors $\mathbf{h_s}$, $\mathbf{r_s}$, and $\mathbf{t_s}$ based on the score function of TransE, formulated as follows:$$f(\mathbf{h_s, r_s, t_s})=||\mathbf{h_s+r_s-t_s}||^2_2 \eqno{(1)} $$We also adopt the following margin-based ranking criterion in our learning:$$f_{loss} = \sum_{(h, r, t)\in \mathcal{G}}\sum_{(h', r, t')\in \mathcal{G}_c}[\gamma + f(\mathbf{h_s}, \mathbf{r_s}, \mathbf{t_s}) - f(\mathbf{h'_s}, \mathbf{r_s}, \mathbf{t'_s})]_+\eqno{(2)}$$,where $\mathcal{G}_c$ is a set of corrupted triples by replacing head entities or tail entities of the triples existing in $\mathcal{G}$ by other random entities in $\mathcal{E}$, $\gamma$ denotes the margin hyperparameter, and $[\cdot]_+$ denotes a Rectified Linear Unit (ReLU). Finally, the structural embedding vectors are learnt by minimizing Equ. (2).

It is worth mentioning that, our model does not constrain the method for structural embedding learning. More sophisticated methods can also be adopted easily.

\subsection{Literal embedding computing}
In this module, we learn the literal embedding vectors of entities and relations to represent the literal information of the educational KG. Specifically, for each triple $(h,r,t) \in \mathcal{G}$, we represent the literals of entities and the relation (i.e., $l_h$, $l_r$, and $l_t$) by learning literal embedding vectors denoted as $\mathbf{h_l}$, $\mathbf{r_l}$, and $\mathbf{t_l}$. 

We employ a state-of-the-art pre-trained language model, i.e., BERT~\cite{devlin2018bert}, to perform this process. BERT includes a multi-layer bidirectional Transformer encoder which takes three embeddings as the input, i.e., token embeddings, segment embeddings, and position embeddings~\cite{devlin2018bert}. We respectively add [CLS] and [SEP] to the head and tail of each literal, tokenize the literal, and encode the literal to vocabulary indices as the token embedding. Since we do not consider the problem of next sentence prediction, the elements of segment embeddings are all set to zero. For position embeddings, we directly adopt the pre-trained model “bert-base-uncased” of Transformers$\footnote{https://github.com/huggingface/transformers\#quick-tour}$ . The pre-trained encoder of BERT computes the hidden vector of each input token, including [CLS] and [SEP]. As a common method, we take the computed hidden vector of [CLS] as the embedding vector of the input literal, and assign it as the literal embedding vector of the corresponding entity or relation.

\subsection{Joint embedding learning}

After the above two modules, for a triple $(h,r,t) \in \mathcal{G}$, we have learnt structural embedding vectors, i.e., $\mathbf{h_s}$, $\mathbf{r_s}$ and $\mathbf{t_s}$, and literal embedding vectors, i.e., $\mathbf{h_l}$, $\mathbf{r_l}$ and $\mathbf{t_l}$. The last step of our model is to combine the structural and literal embedding vectors into joint embedding vectors, i.e., $\mathbf{h_j}$, $\mathbf{r_j}$, and $\mathbf{t_j}$. Since we employ TransE to learn the structural embedding vectors, we continue to adopt the translation mechanism of TransE in the joint learning process.

Firstly, for the combination of structural and literal embedding vectors, we leverage the Gated Recurrent Unit (GRU) proposed by Cho et al.~\cite{cho2014learning}. GRU is similar to LSTM~\cite{hochreiter1997long} but simpler since it has fewer parameters via combining the forget gate and the input gate into a single update gate. Specifically, we train three separate single-layer GRUs for head entities, relations, and tail entities, as illustrated in Fig. \ref{fig:2}. Structural embedding vectors are set as the input vectors of GRUs, and literal embedding vectors are set as the initial hidden states of GRUs. Taking the head entity   as an example, given its structural and literal embedding vectors $\mathbf{h_s}$ and $\mathbf{h_l}$, the computing of its joint embedding vector $\mathbf{h_j}$ is formulated as follows:
\begin{equation}
	\mathbf{r} = \sigma(\mathbf{W_{sr}} \mathbf{h_s} + \mathbf{b_{sr}} + \mathbf{W_{lr}\mathbf{h_l} + b_{lr}} \tag{3}) 
\end{equation}
\begin{equation}
	\mathbf{z} = \sigma(\mathbf{W_{sz}} \mathbf{h_s} + \mathbf{b_{sz}} + \mathbf{W_{lz}} \mathbf{h_l} + \mathbf{b_{lz}} \tag{4})
\end{equation}
\begin{equation}
	\mathbf{n} = \tanh(\mathbf{W_{sn}} \mathbf{h_s} + \mathbf{b_{sn}} + \mathbf{r} * (\mathbf{W_{ln}} \mathbf{h_l} + \mathbf{b_{ln}})  \tag{5})
\end{equation}
\begin{equation}
	\mathbf{h_j} = (1 - \mathbf{z}) * \mathbf{n} + \mathbf{z} * \mathbf{h_l} \tag{6})
\end{equation}
where $\mathbf{W_{sr}}$, $ \mathbf{W_{sz}}$, and $ \mathbf{W_{sn}}$ are input-hidden weighs for $\mathbf{h_s}$, $ \mathbf{W_{lr}}$, $ \mathbf{W_{lz}}$, and $ \mathbf{W_{ln}}$ are hidden-hidden weights for $\mathbf{h_l}$, $\mathbf{b_{sr}}$, $\mathbf{b_{sz}}$, $ \mathbf{b_{sn}}$, $ \mathbf{b_{lr}}$, $ \mathbf{b_{lz}}$, and $ \mathbf{b_{ln}}$ are bias vectors. In the above equations, $\mathbf{z}$ is the update gate vector, $\mathbf{r}$ is the reset gate vector, $\mathbf{n}$ is the new gate, and the operator $*$ denotes the Hadamard product.

Then, we adopt Equ. (1) as the score function of the joint learning, which is formulated as $f(\mathbf{h_j, r_j, t_j})=||\mathbf{h_j+r_j-t_j}||^2_2$. Analogically, we generate corrupt triples and employ Equ. (2) as the loss function. We minimize the loss function over the parameters of the three GRUs and the input structural and literal embedding vectors to learn the target joint embedding representations. 

Another thing we would like to discuss is that, the combination of structural and literal embedding vectors could also be performed through other passways. Here we provide one of our further ideas which is to regard the given triple as a sequence and utilize GRU or LSTM to compute the hidden representation of the tail as the representation of the triple. Then, we can train a neural network to compute the score of this equation. In the future, we will investigate this part in depth.

\section{EXPERIMENTS}
In this section, we analyze the performance of our proposed model on the task of link prediction~\cite{bordes2013translating} over several datasets and compare our model against several baselines.

All experiments were implemented in Python and were conducted on a Linux server with 4 GeForce GTX 1080 GPUs, and Intel Core i7-6900K 3.20GHz 16-core processors with 128 GB memory running Ubuntu 16.04.6 LTS.

\subsection{Educational KG construction for experiments}

The construction of our educational KGs contains two parts. Firstly, we extract two sub-graphs from a large-scale educational KG, i.e., Knowledge Forest$\footnote{http://yotta.xjtushilei.com/data-management-new/module/construct/index.html}$ . As we have introduced in Section 3.1, Knowledge Forest is a KG which includes teaching knowledge of several courses, e.g., data structure, Java, mathematics, biology, etc. Since Knowledge Forest is constructed automatically, and it is very large in size, we only focus on two courses, i.e., Java and data structure, to avoid the inefficiency and any potential quality issues. Two sub-graphs corresponding to the two courses are called DS-KF and Java-KF, where KF stands for Knowledge Forest. In the original Knowledge Forest, there are 193 topics, 35,076 knowledge fragments, and 247 learning dependencies about the course data structure, and there are 173 topics, 50,507 knowledge fragments, and 752 learning dependencies about the course Java. As analyzed in Section 3.1, we treat both topics and knowledge fragments as entities, while learning dependencies are considered as relations. Secondly, for a more comprehensive inspection, we also follow the construction method proposed by Zheng et al.~\cite{zheng2019knowledgeforest} to extract entities, relations, and literals from Wikipedia which are relevant to data mining, computer network, and data structure. And we call this new dataset as CS-Wiki. The statistics of the three datasets are provided in Table \ref{tab:1}.

\begin{table}[t]
	
	\caption{Summary statistics of our educational KGs.}
	\label{tab:1}
	\begin{tabular}{ c | c | c | c }
		\hline
		& CS-Wiki & Java-KF & DS-KF \\
		\hline
		\#Entities & 2022 & 799 & 1164 \\
		
		\#Relations & 1661 & 225 & 277  \\
		
		\#Triples for train & 1789 & 697 & 862 \\
		
		\#Triples for validation & 222 & 96 & 91 \\
		
		\#Triples for test & 231 & 79 & 117 \\
		\hline
	\end{tabular}
\end{table}

\subsection{Link prediction over Knowledge Forest}

The link prediction task is to predict the missing head/tail entities of incomplete test triples. All entities of the given KG are regarded as candidates. For each test triple, we generate a set of candidate triples by replacing the missing entities with candidate entities. The candidate triple set is evaluated by the score function of TransE, i.e., Equ. (1). Following the set-up of TransE~\cite{bordes2013translating}, we adopt two evaluation metrics: (1) Mean Rank, which indicates the mean rank of all correct predictions; and (2) Hits@10, which is the proportion of correct predictions ranked in top-10. A higher Hits@10 and a lower Mean Rank mean a better performance.

\begin{table}[H]
	\caption{Link prediction results on educational KGs.}
	\label{tab:2}
	\begin{tabular}{ccccc}
		\hline
		\multicolumn{5}{c}{CS-Wiki}                                                                                                                      \\ \hline
		\multicolumn{1}{l|}{}                                 & \multicolumn{2}{c|}{Mean Rank (MR)}                & \multicolumn{2}{c}{Hits@10 (100\%)} \\ \hline
		\multicolumn{1}{c|}{Algorithm}                        & Raw           & \multicolumn{1}{c|}{Filtered}      & Raw              & Filtered         \\ \hline
		\multicolumn{1}{c|}{TransE (Bordes et al., 2013)}     & 1088.1        & \multicolumn{1}{c|}{1088.1}        & 14.7             & 14.7             \\ \hline
		\multicolumn{1}{c|}{DistMult (Yang et al., 2014)}     & -             & \multicolumn{1}{c|}{997.3}         & -                & 14.9             \\ \hline
		\multicolumn{1}{c|}{ComplEx (Trouillon et al., 2016)} & -             & \multicolumn{1}{c|}{998.6}         & -                & 14.9             \\ \hline
		\multicolumn{1}{c|}{SimplE (Kazemi, \& Poole, 2018)}  & 1009.3        & \multicolumn{1}{c|}{1009.3}        & 15.8             & 15.8             \\ \hline
		\multicolumn{1}{c|}{TuckER (Balažević et al., 2019)}  & -             & \multicolumn{1}{c|}{919.3}         & -                & 15.8             \\ \hline
		\multicolumn{1}{c|}{RotatE (Sun et al., 2019)}        & -             & \multicolumn{1}{c|}{1228.9}        & -                & 14.5             \\ \hline
		\multicolumn{1}{c|}{Our method}                       & \textbf{68.8} & \multicolumn{1}{c|}{\textbf{68.7}} & \textbf{32}      & \textbf{32}      \\ \hline
	\end{tabular}
\end{table}

\begin{table}[H]
	
	\begin{tabular}{ccccc}
		\hline
		\multicolumn{5}{c}{Java-KF}                                                                                                                      \\ \hline
		\multicolumn{1}{l|}{}                                 & \multicolumn{2}{c|}{Mean Rank (MR)}                & \multicolumn{2}{c}{Hits@10 (100\%)} \\ \hline
		\multicolumn{1}{c|}{Algorithm}                        & Raw           & \multicolumn{1}{c|}{Filtered}      & Raw              & Filtered         \\ \hline
		\multicolumn{1}{c|}{TransE (Bordes et al., 2013)}     & 443.9         & \multicolumn{1}{c|}{441.5}         & 38.6             & 38.6             \\ \hline
		\multicolumn{1}{c|}{DistMult (Yang et al., 2014)}     & -             & \multicolumn{1}{c|}{606.7}         & -                & 18.9             \\ \hline
		\multicolumn{1}{c|}{ComplEx (Trouillon et al., 2016)} & -             & \multicolumn{1}{c|}{598.2}         & -                & 18.9             \\ \hline
		\multicolumn{1}{c|}{SimplE (Kazemi, \& Poole, 2018)}  & 458.7         & \multicolumn{1}{c|}{456.1}         & 22.2             & 22.2             \\ \hline
		\multicolumn{1}{c|}{TuckER (Balažević et al., 2019)}  & -             & \multicolumn{1}{c|}{\textbf{74.17}}         & -                & \textbf{69.6}             \\ \hline
		\multicolumn{1}{c|}{RotatE (Sun et al., 2019)}        & -             & \multicolumn{1}{c|}{453.7}         & -                & 19.0             \\ \hline
		\multicolumn{1}{c|}{Our method}                       & \textbf{106.9}         & \multicolumn{1}{c|}{105.2}         & \textbf{44.9}    & 49.4  \\ \hline
	\end{tabular}
\end{table}

\begin{table}[H]
	
	\begin{tabular}{ccccc}
		\hline
		\multicolumn{5}{c}{DS-KF}                                                                                                                      \\ \hline
		\multicolumn{1}{l|}{}                                 & \multicolumn{2}{c|}{Mean Rank (MR)}                & \multicolumn{2}{c}{Hits@10 (100\%)} \\ \hline
		\multicolumn{1}{c|}{Algorithm}                        & Raw           & \multicolumn{1}{c|}{Filtered}      & Raw              & Filtered         \\ \hline
		\multicolumn{1}{c|}{TransE (Bordes et al., 2013)}     & 396.2        & \multicolumn{1}{c|}{394.3}         & 31.6             & 32.1             \\ \hline
		\multicolumn{1}{c|}{DistMult (Yang et al., 2014)}     & -             & \multicolumn{1}{c|}{804.3}         & -                & 17.5             \\ \hline
		\multicolumn{1}{c|}{ComplEx (Trouillon et al., 2016)} & -             & \multicolumn{1}{c|}{780.44}        & -                & 17.5             \\ \hline
		\multicolumn{1}{c|}{SimplE (Kazemi, \& Poole, 2018)}  & 549.6        & \multicolumn{1}{c|}{547.6}         & 19.6             & 20.1             \\ \hline
		\multicolumn{1}{c|}{TuckER (Balažević et al., 2019)}  & -             & \multicolumn{1}{c|}{\textbf{46.12}}         & -                &\textbf{71.4}              \\ \hline
		\multicolumn{1}{c|}{RotatE (Sun et al., 2019)}        & -             & \multicolumn{1}{c|}{544.7}         & -                & 17.1             \\ \hline
		\multicolumn{1}{c|}{Our method}                       & \textbf{216.2} & \multicolumn{1}{c|}{214.5} & \textbf{36.8}      & 37.6     \\ \hline
	\end{tabular}
\end{table}

Specifically, taking the head prediction as an example, for each test triple $(h,r,t) $, we remove the head entity $h $ to get the incomplete test triple $(?h,r,t) $ and compose the candidate triple set $\mathcal{T}_c = \{(h_c,r,t)|h_c \in \mathcal{E}\}$, where $h_c$ is a candidate entity in $\mathcal{G}$. Then we rank the candidate triples of $\mathcal{T}_c $ in descending order of the cost scores calculated by Equ. (1). It is worth mentioning that, except the target entity $h_{target} $, among the predicted results, there may also exist entities $h_{correct} \in \mathcal{E}$ which satisfy that $(h_{correct},r,t) \in KG$. The above evaluation may rank these triples higher than the test triple since they are also correct. Obviously, it will influence the evaluation. Hence, before ranking we may filter out these corrupted triples which have appeared in the KG. The evaluation setting without filtering is named as “Raw” while the filtered one as “Filter”.

For TransE, we re-implemented it in PyTorch. As for the DistMult, ComplEx, RotatE, we directly utilized the re-implementation by Sun et al.$\footnote{https://github.com/DeepGraphLearning/KnowledgeGraphEmbedding}$~\cite{sun2019rotate}. And we used the code of SimplE$\footnote{https://github.com/baharefatemi/SimplE}$ released by themselves~\cite{kazemi2018simple}. For TransE and our method, we set learning rate $lr=0.0005 $, mini-batch size $b = 256 $, margin $m = 1 $, embedding dimension $k = 50 $ and standard L2 regularization $n=2  $ for both structural and joint embedding learnings. As for the other baselines, we kept the default parameters of the implementations their released. The evaluation results of link prediction on our constructed educational KGs are reported in Table \ref{tab:2}. Since some baseline implementations only provide the codes for computing filtered results, we use dashes to replace the missing values of raw results.

We have two observations from Table \ref{tab:2}:

\begin{itemize}
	\vspace{-0.2cm}
	\item[1.]On CS-Wiki, our method outperforms all baseline models in both Hits@10 and Mean Rank significantly which proves the superiority of our model over other baselines on educational KGs. This is reasonable since they ignore the literal information of educational KGs which is quite important.
	
	\item[2.]On other two datasets, i.e., Java-KF and DS-KF, our model is both ranked in the second place after TuckER. However, our model is still competitive compared to the other baselines. As for why TuckER outperforms our model, we summarize three possible reasons: Firstly, there are more low-quality literals in Java-KF and DS-KF than in CS-Wiki, which may have negative impacts on our literal embedding learnings. For example, we observed that a large portion of the literals of Java-KF are Java codes which are apparently not valuable for our embedding learning. Secondly, we shared the same parameters on the three datasets, however, as illustrated in Table. \ref{tab:1}, the three datasets have different topological features and they also include different literals. Therefore, the learning parameters may need to be further tuned. Finally, as we discussed during the model introduction, both the structural embedding learning and the joint learning modules can be implemented more sophisticatedly. Our current model still has an enormous improvement space.
	
\end{itemize}

\subsection{Effectiveness evaluation over common benchmarks}
Although our embedding model is specially designed for educational KGs which have rich literals, to fully scrutinize our model, we compare it with the embedding model we adopted, i.e., TransE, on three common benchmark KGs, which are WN18~\cite{bordes2014semantic}, FB15k~\cite{bordes2013translating}, and FB15K-237~\cite{toutanova2015representing}. For our model, we set learning rate $lr=0.0005 $, mini-batch size $b = 256 $, margin $m = 1 $, embedding dimension $k = 50 $ and standard L2 regularization $n=2  $ for WN18, set learning rate $lr=0.0005 $, mini-batch size $b = 128 $, margin $m = 1 $, embedding dimension $k = 50 $ and standard L2 regularization $n=2  $ for FB15k and FB15k-237. And for TransE, we set learning rate $lr=0.0005 $, mini-batch size $b = 256 $, margin $m = 1 $, embedding dimension $k = 50 $ and standard L2 regularization $n=2  $ for WN18, set learning rate $n=2  $, mini-batch size $b = 512 $, margin $m = 1 $, embedding dimension $k = 50 $ and standard L2 regularization $n=2  $ for FB15k and FB15k-237. The results of the Mean Rank (Raw/Filter) are reported in Table \ref{tab:3}. We can observe that our model outperforms TransE on both WN18 and FB15k-237. The results demonstrate that the model is not only suitable for education KGs, but also effective on common KGs.

\begin{table}[H]
	\caption{Link prediction results on different methods.}
	\label{tab:3}
	\begin{tabular}{c|cc|cc|ll}
		\hline
		\multicolumn{1}{l|}{}                                                    & \multicolumn{2}{c|}{WN18}      & \multicolumn{2}{c|}{FB15k}         & \multicolumn{2}{l}{FB15K-237}  \\ \hline
		\multirow{2}{*}{Algorithm}                                               & \multicolumn{2}{c|}{Mean Rank} & \multicolumn{2}{c|}{RawMean Rank} & \multicolumn{2}{l}{Mean Rank}   \\ \cline{2-7} 
		& Raw            & Filter        & Raw             & Filter          & Raw            & Filter         \\ \hline
		\begin{tabular}[c]{@{}c@{}}TransE (Bordes   et al., 2013)\end{tabular} & 258            & 247           & \textbf{258}    & \textbf{160}    & 398            & 287            \\
		Our method                                                               & \textbf{235}   & \textbf{225}  & 291             & 205             & \textbf{351.5} & \textbf{217.3} \\ \hline
	\end{tabular}
\end{table}

\section{CONCLUSIONS}

In this paper, we focus on the embedding learning task of educational KGs. The feature distinguishing educational KGs, e.g., Knowledge Forest, from general KGs, e.g., DBpedia, is the richness of literal information. Different from conventional KG embedding techniques which rely on structural relationships among entities and relations, we propose a novel embedding learning model which learns the joint embedding representations based pre-trained structural and literal embedding vectors. The structural embedding vectors are pre-trained by a translation-based embedding model, i.e., TransE. And the literal embedding vectors are pre-trained by a state-of-the-art literal representation model, i.e., BERT. In the joint learning step, we utilize three separate GRUs for head entities, relations, and tail entities, respectively, to combine the structural and literal embedding vectors. And the translation mechanism of TransE is adopted as the loss function of our joint learning step.

We constructed three experimental educational KGs for our experiments, and they could be adopted as benchmark datasets of the future work on educational KGs. The results of our experiments on the constructed educational KGs and common benchmark KGs have demonstrated the effectiveness and superiority of our proposed model. We also noticed that the performance of our model is not the most competitive on several common benchmark KGs. Therefore, in the future, we plan to adopt more improved embedding models which have more sophisticated mechanisms to our model. And we are also trying to include the fine-tuning part of BERT in the joint learning of our model to achieve better performances.

\section{ACKNOWLEDGMENTS}

This work was supported by National Key Research and Development Program of China (2018YFB1004500), National Natural Science Foundation of China (61532015, 61532004, 61672419, and 61672418), Innovative Research Group of the National Natural Science Foundation of China(61721002), Innovation Research Team of Ministry of Education (IRT\_17R86), Project of China Knowledge Centre for Engineering Science and Technology.

\renewcommand\refname{REFERENCES}
\bibliographystyle{splncs04}
\bibliography{191031_yaosiyu_joint-embedding}

\end{document}